\documentclass[final]{camera_ready}

\usepackage{titling}
\pretitle{\begin{center}\Large \bf }
\posttitle{\par\end{center}}
\preauthor{\begin{center}\large\lineskip .5em\begin{tabular}[t]{c}}
\postauthor{\end{tabular}\end{center}}
\predate{\par\begin{center}\large\lineskip .5em}
\postdate{\par\end{center}\vskip .5em\vspace*{12pt}}
\usepackage{times}
\usepackage{epsfig}
\usepackage{graphicx}
\usepackage{amsmath}
\usepackage{amssymb}

\usepackage{multirow}
\usepackage{caption}
\usepackage{subcaption}

\usepackage[pagebackref=true,breaklinks=true,colorlinks,bookmarks=false]{hyperref}

\def\cvprPaperID{4720} 
\def\confYear{CVPR 2021}

\def\oursfull{Sparse Signal Superdensity}
\def\ours{$S^{3}$}
\def\ourmoduleafull{\textit{sparse signal expansion}}
\def\ourmodulebfull{\textit{confidence weighting}}

\begin{document}

\title{\ours{}: Learnable \oursfull{} for Guided Depth Estimation}

\author{Yu-Kai Huang \and Yueh-Cheng Liu \and Tsung-Han Wu \and Hung-Ting Su \and Yu-Cheng Chang \and Tsung-Lin Tsou \and Yu-An Wang \and Winston H. Hsu}
\date{National Taiwan University}

\maketitle

\begin{abstract}
Dense depth estimation plays a key role in multiple applications such as robotics, 3D reconstruction, and augmented reality.
While sparse signal, e.g., LiDAR and Radar, has been leveraged as guidance for enhancing dense depth estimation, the improvement is limited due to its low density and imbalanced distribution.
To maximize the utility from the sparse source, we propose \emph{\oursfull{}} (\ours) technique, which expands the depth value from sparse cues while estimating the confidence of expanded region.
The proposed \ours{} can be applied to various guided depth estimation approaches and trained end-to-end at different stages, including input, cost volume and output.
Extensive experiments demonstrate the effectiveness, robustness, and flexibility of the \ours{} technique on LiDAR and Radar signal. 
\end{abstract}

\section{Introduction} \label{sec:introduction}

\begin{figure}
    \centering
    \begin{subfigure}[b]{\linewidth}
        \centering
        \includegraphics[width=\linewidth]{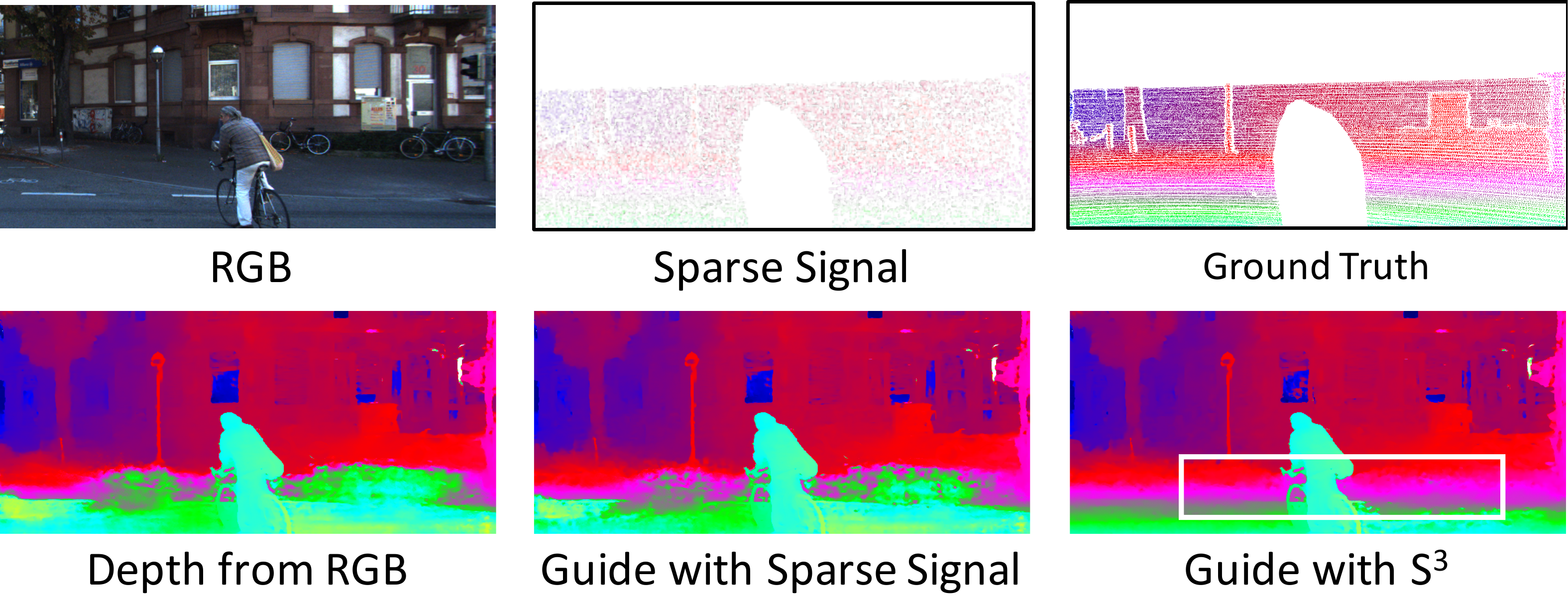}
        \caption{Low Density Problem.}
        \label{fig:figure_1_a}
    \end{subfigure}
    \begin{subfigure}[b]{\linewidth}
        \centering
        \includegraphics[width=\linewidth]{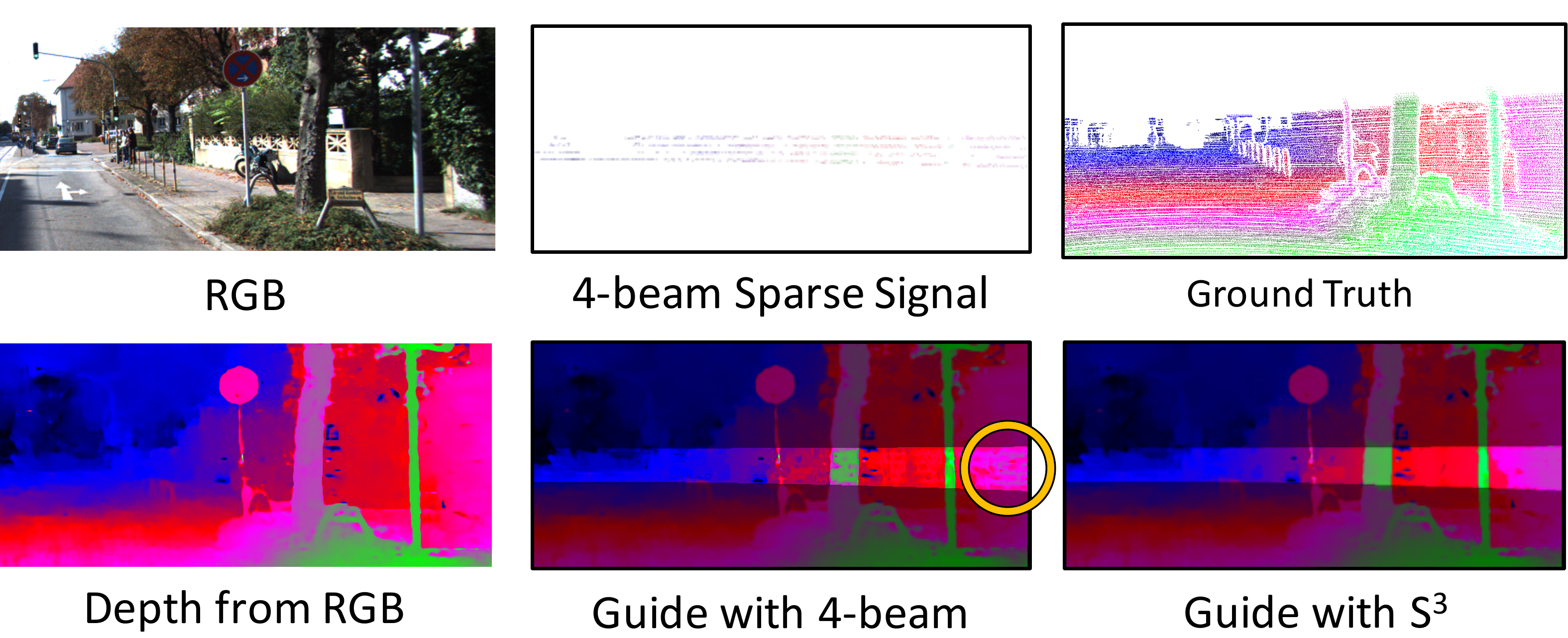}
        \caption{Imbalanced Signal Distribution.}
        \label{fig:figure_1_b}
    \end{subfigure}
    
    \caption{\textbf{Major problems of sparse depth signal.} (a) The network tends to ignore the low density hint if depth from RGB is hugely different from sparse signal. (b) Imbalanced signal distribution would make the guidance to be not equally distributed. We can observe the trace of the 4 scanning lines of LiDAR (yellow circle). Our proposed \ours{} method successfully overcome the two problems. The noise in the top example is removed, and the guided result in the bottom example is smoother and closer to the ground truth. Best viewed in zoomed digital.}
    \label{fig:figure 1}
\end{figure}

Dense depth estimation is crucial in the field of 3D reconstruction \cite{geiger2011stereoscan}, 3D object detection \cite{wang2019pseudo,you2020pseudo}, and robotic vision \cite{marapane1989region,nalpantidis2010stereo}. 
Many works have proposed to estimate depth from RGB images or stereo pairs. Yet, the stereo estimation could be unreliable on homogeneous planes, large illumination changes, and repetitive textures~\cite{shivakumar2019real,wang20193d}; while monocular depth estimation is an ill-posed problem~\cite{eigen2014depth} and inherently ambiguous and unreliable~\cite{lee2019monocular,mal2018sparse}.
To attain a higher level of robustness and accuracy, modern solutions commonly leverage raw sparse signal, such as  LiDAR~\cite{ahmad2020extensible,qiu2019deeplidar,mal2018sparse} and Radar~\cite{chadwick2019distant,nobis2019deep}, to improve depth estimation results or object detection for the challenging outdoor scenes, termed \textit{guidance} in this paper. 

Despite the success of those sparse-guidance methods, however, we still find two big problems with sparse signal. First, raw sparse signal can be ignored by networks when it is largely different from depth predicted with RGB (shown in Figure~\ref{fig:figure_1_a}). This situation stems from the low density property of the sparse signal, which is a common problem in many large-scale dataset. For example, KITTI dataset~\cite{geiger2013vision} wraps up an average density of $4.0\%$ and nuScenes dataset~\cite{nuscenes2019} has an average of less than $50$ Radar points over a $900 \times 1600$ image. Actually, the guidance module tends to ignore the accurate but sparse signals when they strongly disagree with the original prediction.

Furthermore, imbalance guidance is also the main problem. As shown in Figure~\ref{fig:figure_1_b}, the algorithms only focus on the small region with high signal density while barely correct the low-density region between scanning lines and cause non-smoothing result. However, these low-density parts neither implicate less importance nor less confidence. In fact, there could be important objects like cars at these parts, and the imbalanced guidance stems from the uneven signal distribution of sensing devices in space. For example, LiDAR signals are mostly localized on the scanning lines with the same polar angles in the spherical coordinate, and the azimuth resolution of Radar signals is poor~\cite{daniel2018application,sheeny2020300}. As a result, for previous methods that conduct experiments under the assumption of uniformly distributed signal can be unreliable for real-world imbalanced cases.


To tackle the critical \textit{low density} and \textit{imbalanced distribution} problems, we propose a novel framework, \oursfull{} (\ours{}), to enhance the density and mitigate imbalanced sparse signal for guided depth estimation. \ours{} consists of two components: (1) \ourmoduleafull{} (2) \ourmodulebfull{}.
For \ourmoduleafull{}, \ours{} first estimates the expanded area for each sparse signal based on the RGB image, and then assigns appropriate depth value to the expanded region. For \ourmodulebfull{}, \ours{} measures the confidence of the assigned depth to control the amount of influence to the sparse-guidance methods. Our method effectively utilizes \ourmodulebfull{} to increase the density of the sparse signal.

\ours{} framework, implemented with a light-weight network, can be applied to existing sparse-guidance depth estimation methods. For instance, embedding it in existing depth networks and trained in an end-to-end fashion. Losses are developed to allow \ours{} network to learn \ourmoduleafull{} and \ourmodulebfull{} from data either for pretraining purposes or training jointly with depth networks.
We conduct qualitative experiments to show the effectiveness of \ours{} network on LiDAR and Radar guidance methods. The experimental results show that using our proposed \ours{} can solve the \textit{low density} and \textit{imbalanced distribution} problems. Our method can highly increase the utility of the sparse signal and make substantial improvements on four typical sparse-guidance schemes on KITTI~\cite{Geiger2012CVPR,Menze2015CVPR,uhrig2017sparsity} and nuScenes~\cite{nuscenes2019} dataset.

To sum up, our contributions are highlighted as follows,
\begin{itemize}
  \item The first work to point out the defective properties of the sparse signal and the subsequent influence to the depth estimation results.
  \item The novel and general framework \oursfull{} (\ours{}) enhances the density of sparse signal, mitigates the imbalanced distribution problem, and provides extra confidence cues for depth estimation.
  \item \ours{} largely increases the robustness and accuracy on depth estimation tasks using sparse signals, e.g., LiDAR and Radar.
\end{itemize}

\section{Related Work} \label{sec:related_work}
In this section, we will introduce guided depth estimation approaches and review related ideas about signal expansion.

\paragraph{Guided Mono Estimation.} Previous works guide monocular depth estimation networks with external active sensors to address the technically ill-posed problem~\cite{eigen2014depth} and improve performance~\cite{zhang2018deep,huang2019indoor,mal2018sparse,ma2019self,shivakumar2019dfusenet,zhong2019deep,uhrig2017sparsity} known as Depth Completion.
Cheng \etal~\cite{cheng2019learning} fuse the sparse depth as input and propagate the information to the surrounding pixels. Cadena \etal~\cite{cadena2016multi} concatenate the features of the cross-modality data to learn an auto-encoder for completing the partial or noisy depth. Ma and  Karaman~\cite{mal2018sparse} fuse different modalities in the first convolution layer to generate high-resolution depth. The methods aim at completing the depth from sparse depth signal and an image.

\paragraph{Guided Stereo Estimation.} Previous works guide stereo matching results with external sparse signal for better predicted results~\cite{li2020lidar,ahmad2020extensible,park2018high,shivakumar2019real,cheng2019noise}. Stereo matching leverages epipolar geometry to match pixels across image pairs and produce disparity~\cite{zhang1998determining}, which can be transformed to depth by triangulation. PSMNet~\cite{chang2018pyramid} and GANet~\cite{zhang2019ga} are renowned stereo backbones.
Poggi \etal~\cite{poggi2019guided} propose guided techniques on cost volume to alleviate the domain shift. Yet, their method assumes sparse signal to be uniformly distributed, which does not consider imbalanced signal problem.
You \etal~\cite{you2020pseudo} propose a graph-based depth correction algorithm to refine the stereo results in 3D domain with cheap LiDAR sensors. Nonetheless, their algorithm design does not take the imbalanced signal issue into account.
Wang \etal~\cite{wang20193d} propose input fusion and regularize batch normalization conditioning on LiDAR signal.
The above methods utilize the raw sparse signal for guidance or correction, which puts little emphasis on the inherent problems of the sparse signal mentioned.

\begin{figure*}[ht]
    \centering
    \includegraphics[width=\linewidth]{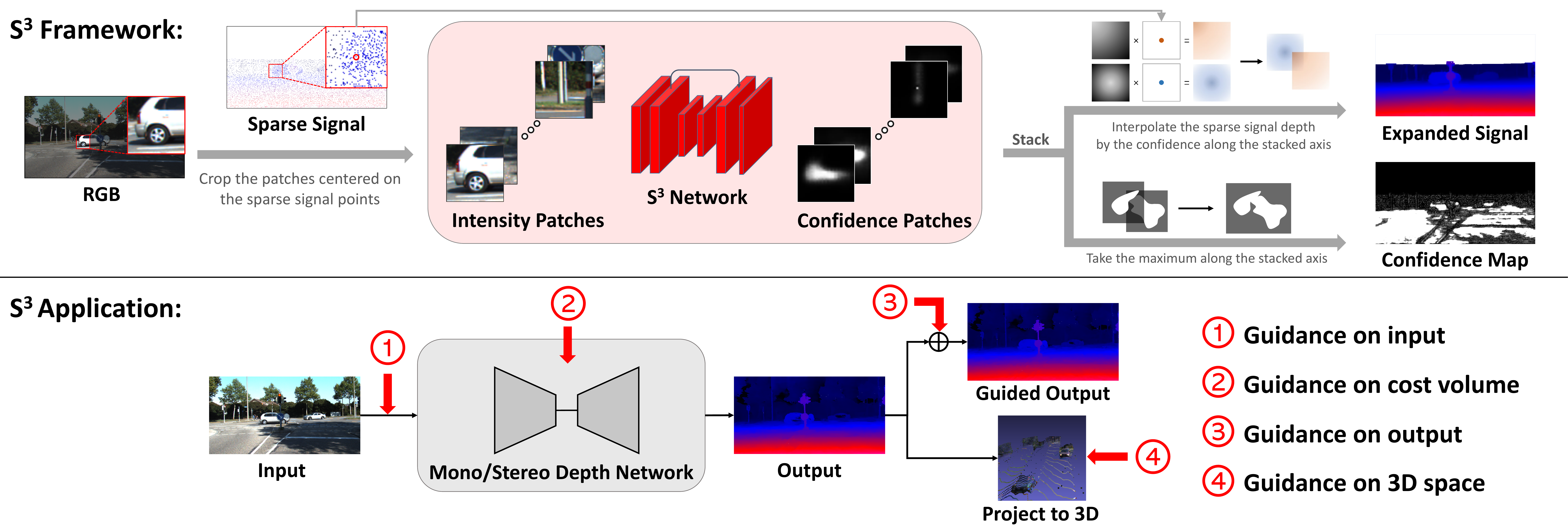}
    \caption{\textbf{\oursfull{} (\ours{}) overview.} The top pipeline illustrates the details of \ours{} framework to expand sparse signal and generate the final expanded depth and confidence map (Section~\ref{subsec:LS3}). The bottom demonstrates the application of our module to guide on different stages of depth estimation (Section~\ref{sec:application_of_S3}).}
    \label{fig:method_LS3}
\end{figure*}

\paragraph{Signal Expansion.} The expansion idea has shown in tasks like superpixel segmentation~\cite{achanta2012slic,van2012seeds,yao2015real,rubio2016bass}, depth completion, and depth sampling~\cite{hawe2011dense,liu2015depth,wolff2020super}. Superpixel aggregates pixels with similar semantics, but they do not imply similar depth values. Depth completion and depth sampling complete the sparse depth, but most of the previous works do not measure the confidence of the expanded depth and rely on heavily computational resources.

Shivakumar \etal~\cite{shivakumar2019real} propose \textit{promotion} of the depth signal to the neighboring pixels in the cost volume to improve depth estimation. The incentive to promote the sparse signal is close to our application on cost volume. However, their methods are only applicable to Semi Global Matching~\cite{hirschmuller2005accurate} algorithm.
Furthermore, there are lots of hand-tuned hyper-parameters and assumptions, like promotion with Gaussian, which may not hold for real data.



\section{Method}

\subsection{Intuition of \oursfull{}} \label{subsec:motivation_of_S3}

To solve the issues of \textit{low density} and \textit{imbalanced distribution}, we propose expanding the sparse cues to the neighbor region. Our idea is that neighboring pixels with similar color intensities belong to the same image structure or object and thus have similar depth values.

\begin{figure}[h!]
    \centering
    \includegraphics[width=\linewidth]{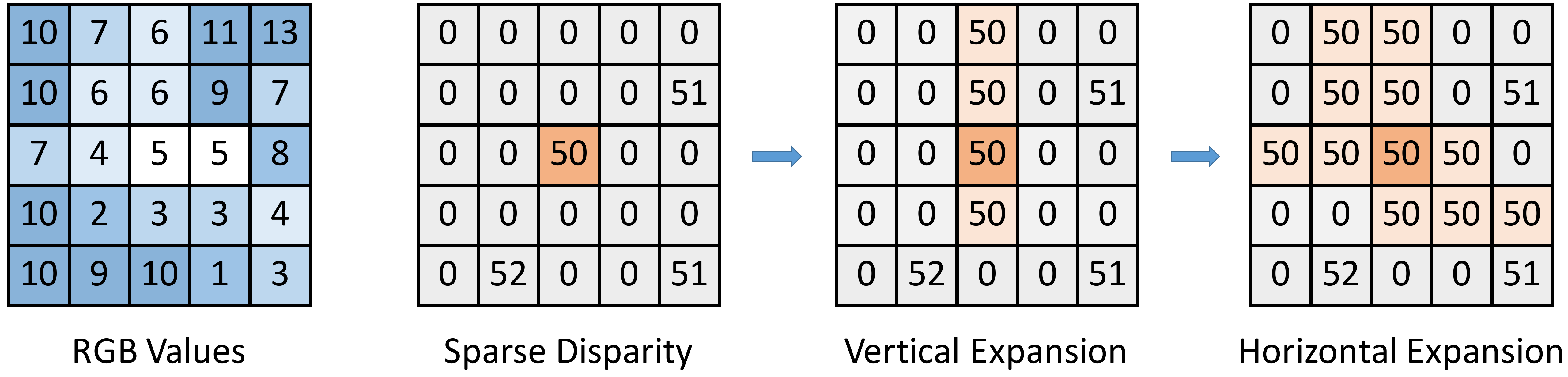}
    \caption{
    \textbf{Intuition for \ourmoduleafull{} by ad-hoc method.} Sparse depth map (right) is expanded according to RGB image (left) presented in one channel here. 
    Zero in the sparse map means no signal. 
    The example expands the center signal according to difference of color intensity with threshold $=2$.}
    \label{fig:fig-2}
\end{figure}

Intuitively, the ad-hoc method is to expand points by color thresholds inspired by cross-based support window method~\cite{zhang2009cross}. To be specific, let $I$, $G$ and $G_{exp}$ be the color intensity map, sparse signal map and expanded map. Given a central pixel $(i, j)$ (the coordinate of the source point), we greedily expand from the central value $G(i, j)$ to its neighbor pixels $(i', j')$ and fill in the expanded pixels $G_{exp}(i', j')$ with $G(i, j)$ as shown in Figure~\ref{fig:fig-2}. The expansion stops until the maximum of color intensity differences is larger than a threshold or the expansion size reaches the limit.

Although the expanded map $G_{exp}$ can substitute the sparse $G$ to perform any guidance techniques in depth estimation, the expanded points may provide false guidance to the estimating process, especially for occlusions or pixels across object boundary. As a result, instead of applying the same level of guidance to all pixels, we provide a confidence map $C$ to measure the reliability of the expanded value in $G_{exp}$ and the level of guidance to apply for depth estimation.

\subsection{Learnable \oursfull{}} \label{subsec:LS3}

We propose leveraging a neural network to learn how to expand  sparse signals and the corresponding confidence with the concept of \ourmoduleafull{} and \ourmodulebfull{} from Section~\ref{subsec:motivation_of_S3}. We expand each sparse signal to a patch by a \ours{} network and aggregate all the expanded patches to form the final output.

To be specific, we predict how confident the sparse depth $G(i, j)$ can expand from the center pixel $(i, j)$ to the neighboring pixel $(i', j')$ with \ours{} network. We set the expansion space to be a square patch of size $2L+1$ for each sparse signal, where $|(i, j) - (i', j')| \leq L$. The input of the $S^3$ network is a crop of the intensity map $I(i-L:i+L, j-L:j+L)$. The output is a confidence patch of the same size and saved in $C_k(i-L:i+L, j-L:j+L) \in [0, 1]$, where $k$ is the index of $k$'th sparse depth signal and $C_k=0$ for other pixels out of the patch. Then, we aggregate the confidence patches to be the expanded depth map $G_{exp}$ by the following interpolation equation:
\begin{equation}
    G_{exp}(i', j') = \frac{\sum_{k \in S_k} C_k(i', j') \cdot G(i_k, j_k)}{\sum_{k \in S_k} C_k(i', j')},
\end{equation}
where $(i_k, j_k)$ is the pixel coordinate of the $k$'th sparse signal and $S_k$ is the set of indices of the sparse signal. The operation means that a pixel with no signal from depth sensors is assigned with an interpolated depth value from its nearby sparse signal values. Consequently, the more confident \ours{} network considers the source signal to be, the more likely the assigned depth value is to be. Finally, we aggregate the confidence maps by taking the maximum among the confidence patches.
\begin{equation}
    C(i', j') = \max_{k \in S_k} C_k(i', j').
\end{equation}
Note that $C(i', j') = 0$ if $(i', j')$ has no expanded signal. $G_{exp}(i', j') = G(i', j')$ and $C(i', j') = 1$ if $(i', j') = (i_k, j_k)$ for a $k \in S_k$.

We formulate a general method to learn \ours{} network along with any depth backbone. Here, the confidence value can act as the weights between the guided depth $G_{exp}$ and the original estimated depth from monocular estimation or stereo matching $D$. That is,
\begin{equation}
    D_{out} = G_{exp} \cdot C + D \cdot (1 - C).
    \label{eq:output_fusion}
\end{equation}
With the depth ground truth $D^*$, the supervised loss on the output depth $D_{out}$ can be formed as $L_{sup} = \|D^* - D_{out}\|$. We also supervise $G_{exp}$ with $D^*$ and add regularization
\begin{equation}
    L_{S^3} = \lambda_1 \cdot C\cdot \|D^* - G_{exp}\| + \lambda_2 \cdot \|C\|.
    \label{eq:reg}
\end{equation}
The first term in Equation~\ref{eq:reg} means the more confident the expanded depth is, the more accurate it should be. The second term prevents excessive confidence for pretraining.
In practice, the gradient of $C$ of the first term is detached, otherwise, $C = 0$ can be a bad local minimum.
The model is trained end-to-end so that the expansion process is learned from data.
The main difference between having and not having \ours{} is that \ours{} increases the density of the sparse signal by providing an additional confidence map to tell the subsequent depth estimation algorithms how reliable the expanded depth is.

\section{Application of \ours{}} \label{sec:application_of_S3}
\ours{} network can learn to expand different modality data, including the most widely used LiDAR and Radar. Furthermore, \ours{} works on both depth and disparity representation, allowing users to use our module in various applications. For instance, disparity is preferred for robotic tasks due to the need to provide higher accuracy in the nearby region~\cite{wang20193d}.

Many works have proposed signal-guidance schemes to enhance depth estimated from RGB as addressed in Section~\ref{sec:introduction} and \ref{sec:related_work}. These methods can be divided into three categories: (1) Guidance on Input and Output (2) Guidance on Cost Volume (3) Guidance on 3D Space. We will introduce how to apply our module for each type of methods (overview in Figure~\ref{fig:method_LS3}) in the following.

\subsection{Guidance on Input and Output} \label{subsec:guidance_on_input_and_output}
For guidance on input, the most intuitive way is to concatenating these external sparse signal as one of the input to the neural network. This strategy is widely used in dense depth estimation domain for either monocular~\cite{zhang2018deep, ma2019self, mal2018sparse} or stereo~\cite{wang20193d} depth estimation. For these approaches, we can simply replace the original raw sparse signal as our expanded signal along with the confidence map. 

For guidance after the output of the depth prediction network, a naive way is to add the accurate but sparse signal to the predicted depth. Similar schemes are used by Chen \etal~\cite{chen2019learning}, called shortcut connection in the paper, and You \etal~\cite{you2020pseudo}, who ignores the sparse signals largely different from stereo results to avoid numerical error and add those signals back to the corrected depth. 
We modify the naive method by interpolation with Equation~\ref{eq:output_fusion} so that more pixels are guided with the expanded $G_{exp}$ and confidence $C$.

\subsection{Guidance on Cost Volume} \label{subsec:guidance_on_cost_volume}
Many practices have tried to modify the cost volume, an intermediate representation of matching relationships between pixels, either guidance with external cues~\cite{poggi2019guided,spyropoulos2014learning,shivakumar2019real} or confidence measure~\cite{poggi2017quantitative} in the field of stereo matching. The cost volume in the stereo network consists of 3D features with geometric and contextual information that allows the subsequent convolution to regress the disparity probability~\cite{kendall2017end,chang2018pyramid,zhang2019ga}.
Here, we take Guided Stereo Matching (GSM)~\cite{poggi2019guided} as an example to explain how \ours{} framework is applied to cost volume. Another example, CCVNorm~\cite{wang20193d}, is presented in the supplementary materials.

GSM~\cite{poggi2019guided} peaks the correlated features of the cost volume suggested from the sparse signal with Gaussian function to provide guidance to the network.
Specifically, let $G \in \mathbb{R}^{H \times W}$ be external sparse but accurate data, $V$ specifies a binary mask whether $G$ has signal on pixel coordinate $(i, j)$, and the cost volume is $CV \in \mathbb{R}^{H \times W \times D_{max} \times F}$, where $D_{max}$ is the max disparity and $F$ is the feature number. 
Given the pixel coordinate $(i, j)$ and disparity value $G(i, j)$ from external cue $G$, they apply Gaussian function
\begin{equation}
    f^{GSM}(i, j, d) = h \cdot e^{-\frac{(d - G(i, j))^2}{2w^2}}
    \label{eq:gsm}
\end{equation}
on the features $CV(i, j, d) \leftarrow ((1 - V(i, j)) \cdot 1 + V(i, j) \cdot f^{GSM}(i, j, d)) \cdot CV(i, j, d)$ of the cost volume, where $h$ and $w$ are hyper-parameters to control the height and width of the Gaussian, $\forall d \in \{0, 1, \cdots, D_{max}-1\}$. The function $f^{GSM}$ enlarges the feature values having positive relation to sparse cues, while suppressing others.

We propose fusing the expanded disparity map $G_{exp}$ and the correspondent confidence map $C$ in a novel approach:
\begin{equation}
    f^{Ours}(i, j, d) = C \cdot \left(h \cdot e^{-\frac{(d - G_{exp}(i, j))^2}{2w^2}}\right) + s.
\end{equation}
The shift range $s$ preserves the minimum feature value when $(d - G_{exp}(i, j))^2$ is large or $C = 0$. When $s$ is positive, value in cost volume $CV(i, j, d)$ will not be suppressed to zero so that the gradient of network would not be blocked during back-propagating. $s$ can be a learnable parameter for training. The confidence value $C$ acts as a switch to control how much guidance should be applied according to the expanded guidance $G_{exp}$.

\begin{figure}[t]
    \centering
    \includegraphics[width=\linewidth]{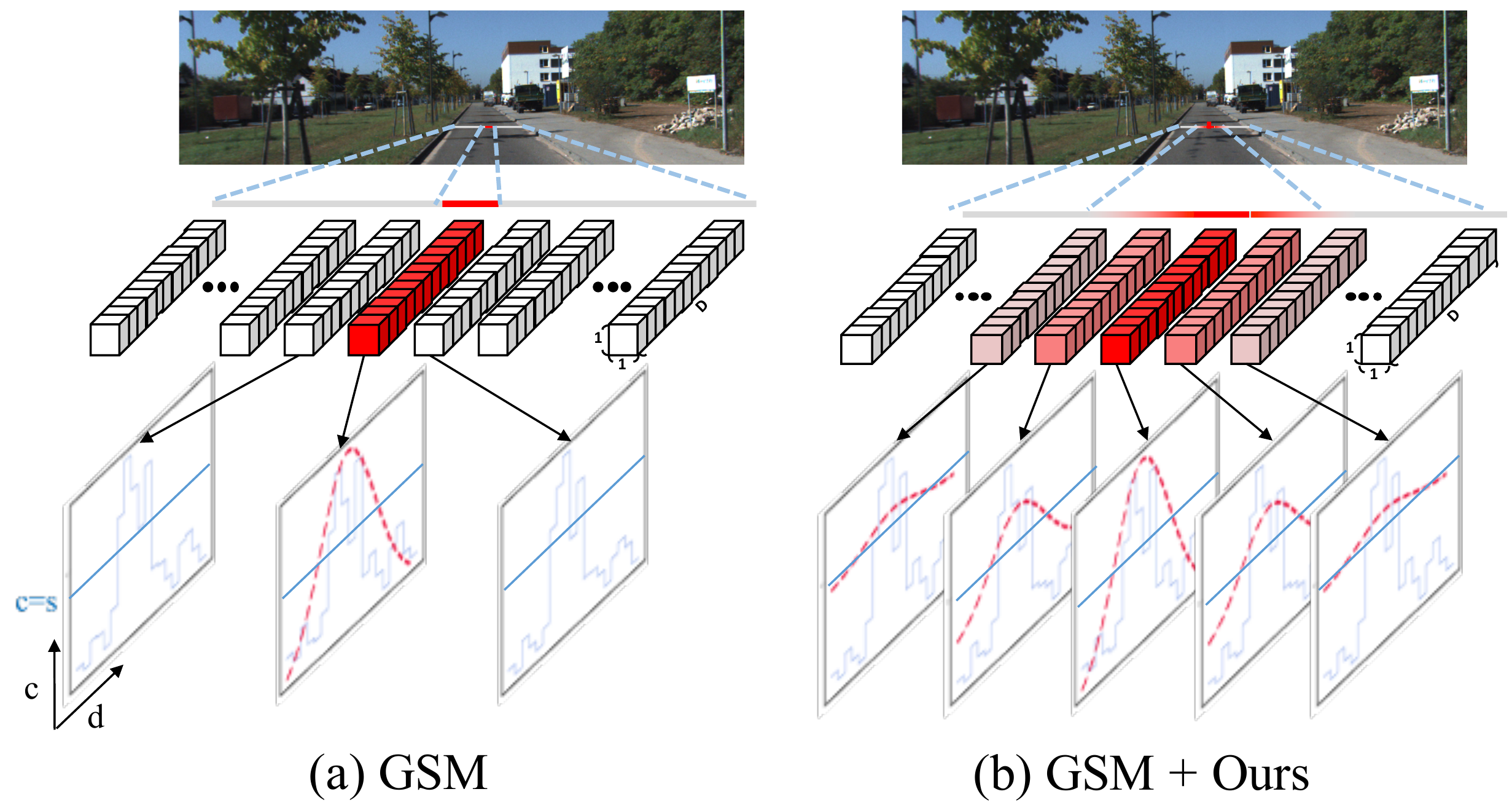}
    \caption{\textbf{Application of \ours{} on cost volume.} We show the slice of the cost volume along a horizontal line. The d-axis denotes the disparity value and c-axis is the cost value. Given a guiding point (red), (a) GSM~\cite{poggi2019guided} guides the features of the point on cost volume. (b) We expand the disparity hint to its neighbors and guide more features with transformed Gaussian based on confidence.}
    \label{fig:gsm_LS3}
\end{figure}

The largest difference between our approach and others are
learnable and confidence-based expansion, which is visualized in Figure~\ref{fig:gsm_LS3}. Additionally, GSM is a subset of ours. Lastly, our module is flexible to apply to other guidance-based approaches like CCVNorm~\cite{wang20193d} on cost volume.

\subsection{Guidance on 3D Space} \label{subsec:guidance_on_3D_space}
In addition to using sparse signal information on input or cost volume, performing sparse signal guidance on 3D space is an intuitive alternative. Take Graph-based Depth Correction (GDC) algorithm proposed by You \etal~\cite{you2020pseudo} as an example, the algorithm first projects the dense depth estimated from monocular or stereo network to 3D space. Then, it forms a neighborhood-relation graph considering depth value via $k$-nearest neighbor.
\begin{equation}
    W = \arg \min_W \|Z - WZ\|^2_2,
    \label{eq:GDC_knn}
\end{equation}
where $Z$ denotes the depth vector, and $W$ denotes the edge weight between two points.
Given the sparse 3D point cloud data, it then corrects the projected points with the relation graph in an optimization manner. 
\begin{equation}
    Z' = \arg \min_{Z'} \|Z' - WZ'\|^2,
    \label{eq:GDC_reconstruction}
\end{equation}
where $Z'_{1:n} = G$. The first $n$ points are set to their correct depth value from the hint of the sparse signals, and the algorithm corrects the rest of points $Z'_{n+1:}$ by minimizing the reconstruction loss. The algorithm corrects the neighbors of the sparse signal points via the relation built from $W$, and the neighbors of the neighbors would also be corrected. The algorithm would propagate the correct depth value via the graph relation for the sparse signals in the long run.

We improve the algorithm with the expanded depth $G_{exp}$ and confidence $C$ in the following approach. Suppose there are $n_e$ expanded points and $m$ points to be corrected, we first built the graph in Equation~\ref{eq:GDC_knn}, and then minimize the reconstruction error considering the confidence.
\begin{multline}
    Z' = \arg \min_{Z'} \|(C'G_{exp}+(I-C')Z') - \\ W(C'G_{exp}+(I-C')Z')\|^2.
\end{multline}
Here $C'\in \mathbb{R}^{(n+n_e+m) \times (n+n_e+m)}$ is a diagonal matrix, where $C'_{kk} = 1$ for $k \in \{1, \cdots, n\}$, $C'_{kk} = C$ for $k \in \{n+1, \cdots, n+n_e\}$, and $C'_{kk} = 0$, otherwise. The modification differs from Equation~\ref{eq:GDC_reconstruction} in that $Z'_{n+1:n+n_e}$ is interpolated to the suggested value $G_{exp}$ with confidence $C$. For $C$ close to $0$, the influence of the guidance value is negligible. For $C$ close to $1$, the guidance value is as confident as the one from sparse signal. Such modification not only allows more points to be corrected by the algorithm, but also takes the magnitude of guidance into consideration.

\section{Experiment}

\subsection{Experimental Setting} \label{exp:experimental_setting}
\paragraph{Dataset.}
We use SceneFlow~\cite{mayer2016large}, KITTI Stereo 2012~\cite{Geiger2012CVPR}, and 2015~\cite{Menze2015CVPR} to conduct experiments for LiDAR sparse signal, and NuScenes v1.0 dataset~\cite{nuscenes2019} for Radar sparse signal.
SceneFlow~\cite{mayer2016large} dataset is a large-scale synthetic stereo dataset mainly for pretraining purpose.
KITTI Stereo 2012~\cite{Geiger2012CVPR} and KITTI Stereo 2015~\cite{Menze2015CVPR} datasets contain stereo and LiDAR data with an application to autonomous driving.
Due to no dense depth ground truth provided on NuScenes, we accumulate consecutive frames of LiDAR signals (5 before and 5 after the frame of interest) for evaluation as KITTI dataset did \cite{Geiger2012CVPR}. 

The sparse signal for KITTI Stereo datasets is obtained according to the original paper. For Guided Stereo Matching (GSM)~\cite{poggi2019guided} experiments, we sub-sample 15\% of pixels from the semi-dense disparity maps. For Graph-based Depth Correction (GDC)~\cite{you2020pseudo} experiments, we obtain the 4-beam LiDAR signal by slicing point clouds into separate lines by an elevation step of $0.4^\circ$.

\paragraph{Training Protocol.}
For GSM~\cite{poggi2019guided}, we pretrain on SceneFlow, fine-tune on the training set of KITTI Stereo 2012, and test on the training set of KITTI Stereo 2015, following the protocols in the original paper. We also fine-tune on KITTI Stereo 2015, and test on KITTI Stereo 2012. 
For GDC~\cite{you2020pseudo}, we use the officially released SceneFlow pretraining from PSMNet~\cite{chang2018pyramid} and fine-tune on the training sets of KITTI Stereo 2012 and 2015, and test on 2015 and 2012, respectively.
For monocular depth estimation on nuScenes dataset, the network is trained supervisedly with L1 loss on LiDAR signal and guided with two algorithms: (1) Guidance on Output in Section~\ref{subsec:guidance_on_input_and_output} (2) GDC in Section~\ref{subsec:guidance_on_3D_space}.

\paragraph{Implementation Detail.} \label{para:implementation_detail}
We implement the proposed methods with PyTorch~\cite{paszke2019pytorch} framework. The architecture of \ours{} network is a light-weight version of U-Net~\cite{ronneberger2015u} structure with patch size $32$ with the last Sigmoid layer to normalize the confidence map. The number of parameters for \ours{} network is $0.7$M and only takes $11\%$ of the depth network like PSMNet~\cite{chang2018pyramid}. The inference time of the module is $0.14$ms per patch for a single thread on one NVIDIA TESTLA V100 GPU with batch size $512$, which can be sped up by parallelism of patch operations.
\ours{} network is pretrained on SceneFlow for $8000$ iterations end-to-end with PSMNet~\cite{chang2018pyramid} optimized with Adam~\cite{kingma2014adam} and $0.001$ learning rate. Following previous works~\cite{chang2018pyramid, zhang2019ga}, we randomly crop $256$ by $512$ for training and pad to full resolution for testing on SceneFlow and KITTI datasets. For nuScenes, we rescale input images to 288 by 512 and train sparse-to-dense~\cite{ma2019self} monocular backbone from scratch for $35$k iterations. Then, the depth is guided by \ours{} network pretrained from SceneFlow.

\paragraph{Evaluation Metric.}
We follow standard metrics to evaluate the results. For disparity maps, we use average pixel error (Avg) and $n$-pixel error rate ($> n$).
The ``Avg" is defined as $\frac{1}{N}\sum{\vert D_{\text{pred}} - D_{\text{gt}} \vert }$
, where $N$ denotes the number of pixels included in valid ground truth disparity map. The ``$> n$" represents the percentage of disparity error that is greater than $n$. We evaluate depth maps with root mean squared (RMS) error, mean absolute relative error (REL), and $\delta_i$. The $\delta_i$ means the percentage of the relative error within a threshold of $1.25^i$.
Except for $\delta_i$, the other metrics are the smaller the better.

\subsection{Guidance Experiment} \label{exp:guidance_experiment}
\subsubsection{Guidance on Input and Output.} \label{exp:guidance_on_input_and_output}

\setlength{\tabcolsep}{0.017\linewidth}{
\begin{table}[]
    \centering
    \begin{tabular}{lcrrrrr} 
         \multirow{2}{*}{Model} & \multirow{2}{*}{\shortstack{Avg Disp\\Error $\downarrow$}} & \multicolumn{5}{c}{$> n$ Disp Error Rate (\%) $\downarrow$}\\ \cline{3-7}
         & & $>1$ & $>2$ & $>3$ & $>4$ & $>5$ \\\hline
         In & 0.891 & 22.72 & 6.12 & 3.02 & 2.09 & 1.63 \\
         \textbf{In + Ours} & \textbf{0.851} & \textbf{21.93} & \textbf{5.98} & \textbf{2.77} & \textbf{1.78} & \textbf{1.34} \\ \hline
         Out & 0.935 & 26.37 & 8.29 & 3.98 & 2.59 & 1.94\\
         \textbf{Out + Ours} & \textbf{0.418} & \textbf{8.90} & \textbf{1.97} & \textbf{1.05} & \textbf{0.73} & \textbf{0.55}\\
    \end{tabular}
    \caption{\textbf{Guidance on Input (In) and Output (Out) Experiments on KITTI Stereo 2015.} (Section~\ref{exp:guidance_on_input_and_output})}
    \label{tab:guidance_on_input_and_output}
\end{table}
}

In Table~\ref{tab:guidance_on_input_and_output}, even though our input guidance simply concatenating the superdensity as input, our approach can still improve upon the guided results with PSMNet.
On the other hand, we contribute the huge gain of our output guidance to the density of the sparse signal, since the only difference is that more pixels are guided by expanded signal. Also, the improvement strengthens our idea that neighboring pixels of the sparse signal have similar depth and are able to be modeled with confidence by the center depth value.

\subsubsection{Guidance on Cost Volume} \label{exp:guidance_on_cost_volume}

\begin{figure*}[ht]
    \centering
    \includegraphics[width=\linewidth]{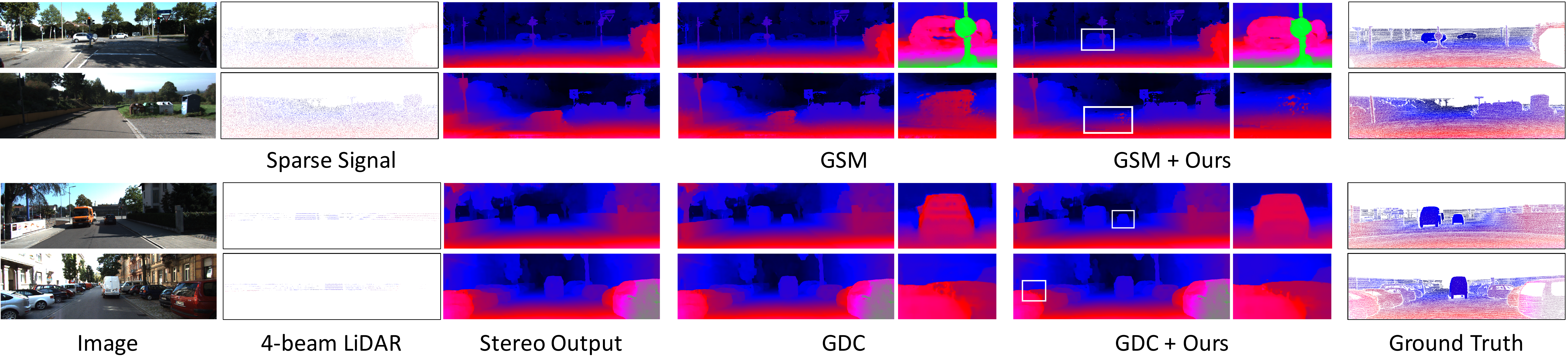}
    \caption{\textbf{Visualization on KITTI Stereo Datasets with Methods GSM~\cite{poggi2019guided} and GDC~\cite{you2020pseudo}.} We show the original depth color map and the zoomed one (visually enhanced) to compare results with (5th column) and without (4th column) our method, which is best viewed in zoomed digital and color. The first row shows that our \ours{} can fix the unreliable matches on the distant cars which is the low density region. The second row demonstrates that the noise from domain shift cannot completely be removed without our method. The third row illustrates that \ours{} reduces the \textit{imbalanced signal distribution} problem, which the scanning lines of LiDAR are obvious in the results of GDC~\cite{you2020pseudo}. The last example shows that the edge of cars are better preserved with our method.}
    \label{fig:visualize}
\end{figure*}

In Table~\ref{tab:gsm}, applying our method in Section~\ref{subsec:guidance_on_cost_volume} on GSM can boost a large gap of performance.
In the visualization results of Figure~\ref{fig:visualize}, GSM does not correct much depth pixel from the stereo output, but it does when applying \ours{}. This tells that the network tends to ignore sparse signal when the density is not high enough, which consents to the \textit{low density} problem and our motivation of solution. Note that we use GANet~\cite{zhang2019ga} as the backbone for no fine-tuning cases because we fail to reproduce GSM  results on PSMNet~\cite{chang2018pyramid}.

\setlength{\tabcolsep}{0.012\linewidth}{
\begin{table}
    \centering
    \begin{tabular}{clcrrrr}
        Dataset & Model & Avg & $>2$ & $>3$ & $>4$ & $>5$\\ \hline \hline
        \multirow{3}{*}{\shortstack{KITTI\\2015}} & GANet~\cite{zhang2019ga} & 1.949 & 20.72 & 12.43 & 8.78 & 6.73 \\
        & + GSM & 1.698 & 15.84 & 9.30 & 6.68 & 5.25 \\
        & \textbf{+ GSM + Ours} & \textbf{1.027} & \textbf{6.65} & \textbf{2.86} & \textbf{1.92} & \textbf{1.51} \\ \hline
        \multirow{3}{*}{\shortstack{KITTI\\2015\\(ft)}} & PSMNet~\cite{chang2018pyramid} & 1.200 & 6.34 & 3.12 & 2.18 & 1.75 \\
        & + GSM & 0.763 & 2.74 & 1.83 & 1.51 & 1.34 \\
        & \textbf{+ GSM + Ours} & \textbf{0.443} & \textbf{1.65} & \textbf{0.96} & \textbf{0.71} & \textbf{0.57}\\ \hline
        
        \multirow{3}{*}{\shortstack{KITTI\\2012}} & GANet~\cite{zhang2019ga} & 1.640 & 17.41 & 11.32 & 8.28 & 6.45 \\
        & + GSM & 1.370 & 12.26 & 7.90 & 5.92 & 4.74 \\
        & \textbf{+ GSM + Ours} & \textbf{0.836} & \textbf{4.70} & \textbf{2.27} & \textbf{1.54} & \textbf{1.18}\\ \hline
        \multirow{3}{*}{\shortstack{KITTI\\2012\\(ft)}} & PSMNet~\cite{chang2018pyramid} & 1.010 & 7.19 & 4.77 & 3.65 & 2.96\\
        & + GSM & 0.526 & 2.68 & 1.76 & 1.34 & 1.10\\
        & \textbf{+ GSM + Ours} & \textbf{0.342} & \textbf{1.37} & \textbf{0.86} & \textbf{0.65} & \textbf{0.52} \\
    \end{tabular}
    \caption{\textbf{Experiments on GSM~\cite{poggi2019guided}.} ``ft" refers to fine-tuning on another KITTI Stereo dataset. (Section~\ref{exp:guidance_on_cost_volume})}
    \label{tab:gsm}
\end{table}
}

\subsubsection{Guidance on 3D Space} \label{exp:guidance_on_3D_space}

\setlength{\tabcolsep}{0.016\linewidth}{
\begin{table*}[]
    \centering
    \begin{tabular}{lccrrrrrccrrrrr}
        \multirow{2}{*}{Model} & \multirow{2}{*}{Fine-tune} & \multicolumn{6}{c}{KITTI Stereo 2012} && \multicolumn{6}{c}{KITTI Stereo 2015}\\ \cline{3-8}\cline{10-15}
        & & Avg & $>1$ & $>2$ & $>3$ & $>4$ & $>5$ && Avg & $>1$ & $>2$ & $>3$ & $>4$ & $>5$ \\\hline \hline
        PSMNet~\cite{chang2018pyramid} & & 8.156 & 89.54 & 78.83 & 68.04 & 57.71 & 48.21 && 8.568 & 86.32 & 73.02 & 60.07 & 48.66 & 38.97 \\
        + GDC & & 7.995 & 84.56 & 74.82 & 65.14 & 55.60 & 46.66 && 8.566 & 83.87 & 71.25 & 58.94 & 47.85 & 38.32 \\
        + GDC + Ours & & 7.776 & 80.32 & 71.27 & 62.34 & 53.45 & 45.01 && 8.479 & 81.84 & 69.60 & 57.78 & 47.03 & 37.74 \\
        \hline
        PSMNet~\cite{chang2018pyramid} & \checkmark & 1.039 & 17.82 & 7.37 & 4.82 & 3.66 & 2.96 && 1.028 & 23.58 & 6.75 & 3.46 & 2.44 & 1.96\\
        + GDC & \checkmark & 0.950 & 15.65 & 6.75 & 4.46 & 3.41 & 2.77 && 0.952 & 21.08 & 6.06 & 3.19 & 2.27 & 1.82 \\
        \textbf{+ GDC + Ours} & \textbf{\checkmark}& \textbf{0.904} & \textbf{14.53} & \textbf{6.31} & \textbf{4.20} & \textbf{3.22} & \textbf{2.62} && \textbf{0.915} & \textbf{20.07} & \textbf{5.76} & \textbf{3.05} & \textbf{2.17} & \textbf{1.75} \\
    \end{tabular}
    \caption{\textbf{Experiments on GDC Algorithm Proposed in Pseudo-LiDAR++~\cite{you2020pseudo}.} (Section~\ref{exp:guidance_on_3D_space})}
    \label{tab:PL++}
\end{table*}
}

In Table~\ref{tab:PL++}, the results show consistent improvement when applying our method in Section~\ref{subsec:guidance_on_3D_space}. The performance gain of GDC is smaller than GSM because the number of points of 4-beam LiDAR is less than the sub-sampled one from GSM. The visualization in the fourth row of Figure~\ref{fig:visualize} illustrates the \textit{imbalanced signal distribution} problem is reduced with our method. The results are presented in the disparity domain, since the Pseudo-LiDAR point cloud~\cite{wang2019pseudo} originates from stereo matching. Also, we evaluate on the task of depth estimation instead of object detection because the focus of this paper is to improve depth estimation results.

\subsection{Radar Guidance} \label{exp:radar_guidance}
We test the effectiveness of our module for Radar signal on nuScenes~\cite{nuscenes2019} dataset, which is one of the first datasets containing Camera, Radar, and LiDAR in diverse scenes and weather conditions. We choose guidance on output and guidance on 3D (GDC~\cite{you2020pseudo}) to improve the prediction of monocular depth estimation shown in Table~\ref{tab:nuScenes}. The improvement of ``GDC + Ours" on LiDAR modality is significant compared to Table~\ref{tab:PL++} because the LiDAR source here is 32-beam instead of 4-beam.
The improvement from Radar modality is minor compared to LiDAR because the number of Radar point cloud is extremely sparse due to small elevation degree. However, with the help of \ours{}, the performance gain can be amplified. The experiment demonstrates the success of our proposed \ours{} framework on both Radar and LiDAR sparse signals.

\setlength{\tabcolsep}{0.01\linewidth}{
\begin{table}
    \centering
    \begin{tabular}{llcccccc}
        Guide & Modal & +Ours & Rel $\downarrow$ & RMS $\downarrow$ & $\delta_1 \uparrow$ & $\delta_2 \uparrow$ & $\delta_3 \uparrow$\\ \hline \hline
        None & - & & 0.161 & 6.79 & 79.71 & 92.05 & 96.15\\ \hline
        Out & Radar & & 0.161 & 6.79 & 79.71 & 92.05 & 96.15\\
        Out & Radar & \checkmark & 0.161 & 6.77 & 79.80 & 92.10 & 96.17\\        
        GDC & Radar & & 0.161 & 6.79 & 79.71 & 92.06 & 96.15\\
        GDC & Radar & \checkmark & \textbf{0.160} & \textbf{6.76} & \textbf{79.96} & \textbf{92.13} & \textbf{96.17}\\ \hline

        Out & LiDAR & & 0.154 & 6.63 & 80.36 & 92.41 & 96.38\\
        Out & LiDAR & \checkmark & 0.090 & 4.59 & 89.63 & 95.74 & 98.05\\
        GDC & LiDAR & & 0.150 & 6.62 & 80.60 & 92.42 & 96.37\\
        GDC & LiDAR & \checkmark & \textbf{0.055} & \textbf{3.64} & \textbf{95.97} & \textbf{97.87} & \textbf{98.79}\\
    \end{tabular}
    \caption{\textbf{Experiments of Radar Signal on NuScenes~\cite{nuscenes2019} Dataset.} ``Out" means guidance on output and GDC is graph-based depth correction~\cite{you2020pseudo}. We demonstrate the ability of our method to gain improvement even on extremely sparse Radar signal. (Section~\ref{exp:radar_guidance})}
    \label{tab:nuScenes}
\end{table}
}

\subsection{Ablation Study} \label{exp:ablation_study}

\paragraph{Effectiveness of Each Component.}
We decompose our module with the expansion part and the confidence part. In Table~\ref{tab:ablation_module}, the main improvement comes from the expansion design, which realizes our arguments that expanding the sparse signal before guidance can improve. When considering the confidence of the expanded signal, \ours{} network is allowed to learn the fine-grained magnitude of influence to the guidance and bring better results.

\setlength{\tabcolsep}{0.017\linewidth}{
\begin{table}
    \centering
    \begin{tabular}{lcccccc}
         Model & Avg & $>1$ & $>2$ & $>3$ & $>4$ & $>5$\\ \hline \hline
         No Correction & 1.010 & 16.87 & 7.19 & 4.77 & 3.65 & 2.96\\
         + Sparse Signal & 0.526 & 6.45 & 2.68 & 1.76 & 1.34 & 1.10\\
         \ \ + Expansion & 0.383 & 4.90 & 1.90 & 1.19 & 0.88 & 0.71\\
         \textbf{\ \ \ \ + Confidence} & \textbf{0.342} & \textbf{3.83} & \textbf{1.37} & \textbf{0.86} & \textbf{0.65} & \textbf{0.52}\\
    \end{tabular}
    \caption{\textbf{Ablation Study of GSM~\cite{poggi2019guided} on KITTI 2012.} The best combination is to add both Expansion and Confidence on Sparse Signal. ``No Correction" refers to the raw stereo output. (Section~\ref{exp:ablation_study})}
    \label{tab:ablation_module}
\end{table}
}

\paragraph{Sparsity Expansion.} 
We discuss on how to expand the sparse signal in Table~\ref{tab:ablation_expansion}. Two baseline models closely related to the idea of expansion are chosen for the experiment: (1) The ad-hoc method mentioned in Section~\ref{subsec:motivation_of_S3}. (2) A superpixel algorithm, SLIC~\cite{achanta2012slic}, which iteratively clusters the neighbor pixels based on color and distance. Confidence weighting is applied to the baselines by considering the inverse distance of the expanded point to the source point, i.e., expanded depth closer to the source has higher confidence.

In Table~\ref{tab:ablation_expansion}, performing expansion on the sparse signal is better than no expansion for no fine-tuning case. This tells that increasing the density of the external signal can help reduce the domain shift problem, where a network is initially trained on a synthetic dataset and tested on real imagery when real data is insufficient. This also meets the goal of improving the overall accuracy without retraining mentioned in GSM~\cite{poggi2019guided}.

For fine-tuning case, simple expansion by color thresholds, like ad-hoc expansion, is worse than no expansion. This implies the stereo network can learn to leverage the sparse signal better than simple expansion techniques. Nevertheless, our proposed \ours{} can jointly learn with the depth network to achieve better results.

The assumption of the confidence weighting for baseline methods may not hold all the time. The expansion of baselines can enlarge the guided field, but it would also provide false guidance to disparity discontinuous areas, where disparity changes sharply. The ablation study results demonstrate the learnable confidence weighting can avoid the ill assumption and improve performance.

\begin{table}
    \centering
    \begin{tabular}{lcc}
         Expansion Model & Avg Error & Avg Error (Fine-tune)\\ \hline \hline
         No Expansion & 1.370 & 0.526 \\
         Ad-hoc Method & 1.155 & 0.582 \\
         SLIC~\cite{achanta2012slic} & 1.027 & 0.489 \\
         \textbf{Ours} & \textbf{0.836} & \textbf{0.342} \\
    \end{tabular}
    \caption{\textbf{Ablation Study of Different Expansion Methods on KITTI 2012 Applied with GSM~\cite{poggi2019guided}.} (Section~\ref{exp:ablation_study})}
    \label{tab:ablation_expansion}
\end{table}

\begin{figure}
    \centering
    \includegraphics[width=\linewidth]{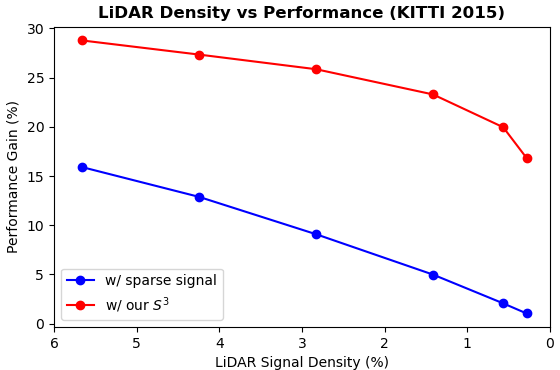}
    \caption{\textbf{Density vs Performance.} The figure stresses the robustness of \ours{} for extremely low signal density.
    }
    \label{fig:density}
\end{figure}

\paragraph{Robustness.}
We also test the robustness of \ours{} by sampling different density of the external signal in Figure~\ref{fig:density}. Surprisingly, our method with merely $0.28\%$ of sparse data beats GSM with $20$ times denser, which strongly supports the idea to increase density of sparse data for guidance.
In addition, our prediction suffers little performance drop until the external cue is extremely sparse, which emphasizes the robustness of \ours{} to work under extreme environment.

\section{Conclusion}
In the paper, we propose \ours{} framework to improve depth estimation results by considering the defective property of sparse signals. Our idea is deployable to existing sparse-guidance methods. Extensive experiments show consistent improvement among guidance approaches, and strengthen the idea that expansion on sparse signal can solve \textit{low density} and \textit{imbalanced distribution} problem. Our \ours{} framework could become an important reference for future exploration on sparse-guidance methods.

\section*{Acknowledgement}
This work was supported in part by the Ministry of Science and Technology, Taiwan, under Grant MOST 110-2634-F-002-026 and FIH Mobile Limited. We benefit from NVIDIA DGX-1 AI Supercomputer and are grateful to the National Center for High-performance Computing.

\clearpage
\setcounter{section}{0}

\title{Supplementary Material:\\\ours{}: Learnable \oursfull{} for Guided Depth Estimation}
\author{}
\date{}
\maketitle

\section{\ours{} Framework Tradeoff}
We discuss the tradeoff between the performance and overhead by dividing the key factors into (1) patch size (2) sample rate, and (3) model size. 

For (1) the patch size cropped by the center of sparse signals, doubling the size would quadruple the tensor memory and inference time, and the performance would improve but converge till the sparse cues are effective enough for a local structure.

For (2), the sample rate is the \% of the sparse signals chosen for expansion by \ours{}, other sparse signals remain the same. Higher sample rate would cover and overlap more expanded region without extra memory but increase the computational cost linearly. We find that 25\% sample rate can effectively reduce the inference time without hurting much performance.

For (3) the model size (altered by number of channels and convs), reducing the model size effectively reduces the memory usage and inference time, but suffers performance drop larger than (2) the sample rate. 

Here we highlight the flexibility to apply our \ours{} framework. If a user prefers real-time usage, then he or she should use a small sample rate. If a user prefers to reduce the memory usage, then he or she should use a small patch size. And if a user wants to reach state-of-the-art performance, then he or she should maximize the sample rate and model size.

\section{More Guidance on Cost Volume}
\subsection{Guidance on Batch Normalization}
Wang \etal~\cite{wang20193d} propose to add guidance to the batch normalization in the cost volume. They leverage Conditional Batch Normalization (CBN) operation to predict the feature-wise affine transformation in dependence on the condition of sparse LiDAR signal $L^s$. In particular, the CBN can be written in the following given a mini-batch of data indexed $i$ and cost volume features $F \in \mathbb{R}^{C \times H \times W \times D}$,
\begin{align}
    & F^{CCVNorm}_{i, c, h, w, d} = \gamma_{i, c, h, w, d} \frac{F_{i, c, h, w, d} - \mathbb{E}_{\beta}[F_{\cdot, c, \cdot, \cdot, \cdot}]}{\sqrt{Var_{\beta}[F_{\cdot, c, \cdot, \cdot, \cdot}] + \epsilon}} + \beta_{i, c, h, w, d}\\
    & \gamma_{i, c, h, w, d} = 
    \begin{cases}
        \phi^g(d) g_{c}(L^s_{i, h, w}) + \psi^g(d), &\text{ if } L^s_{i, h, w} \text{ is valid}\\
        \overline{g}_{c, d}, & \text{ otherwise}
    \end{cases}
    \label{eq:ccvnorm_gamma}\\
    & \beta_{i, c, h, w, d} = 
    \begin{cases}
        \phi^h(d) h_{c}(L^s_{i, h, w}) + \psi^h(d), &\text{ if } L^s_{i, h, w} \text{ is valid}\\
        \overline{h}_{c, d}, & \text{ otherwise}
    \end{cases}.
\end{align}
The $\gamma$ and $\beta$ parameters are conditioned on the sparse source $L^s_{i, h, w}$ if it is valid, otherwise the parameters are reduced to the unconditional ones. $g_c$ and $h_c$ compute the intermediate representations of the sparse signal. $\phi$ and $\psi$ modulate the final normalization parameters based on the intermediate representations. More details are presented in the original paper.

The following we demonstrate how our proposed \ours{} module is applied to the conditional batch normalization. With the expanded disparity $L^{exp}$ and confidence $L^{cnf}$ from \ours{}, we improve the batch normalization process as
\begin{align}
    \gamma^{Ours}_{i, c, h, w, d} =& 
    \begin{cases}
        L^{cnf}_{i, c, h} \cdot \left(\phi^g(d) g_{c}(L^{exp}_{i, h, w}) + \psi^g(d)\right) +\\
        \quad (1 - L^{cnf}_{i, c, h}) \cdot \overline{g}_{c, d}, \text{ if } L^{exp}_{i, h, w} \text{ is valid}\\
        \overline{g}_{c, d}, \text{ otherwise}
    \end{cases}\\
    \beta^{Ours}_{i, c, h, w, d} =& 
    \begin{cases}
        L^{cnf}_{i, c, h} \cdot \left(\phi^h(d) h_{c}(L^{exp}_{i, h, w}) + \psi^h(d)\right) + \\
        \quad (1 - L^{cnf}_{i, c, h}) \cdot \overline{h}_{c, d}, \text{ if } L^{exp}_{i, h, w} \text{ is valid}\\
        \overline{h}_{c, d}, \text{ otherwise}
    \end{cases}.
\end{align}
The intuitive explanation for $\gamma^{Ours}$ is that we interpolate the valid value and invalid one of $\gamma$ in Equation~\ref{eq:ccvnorm_gamma} by the expanded confidence $L^{cnf}$ if $L^{exp}$ is valid.

\subsection{Experiments on Batch Normalization}
We follow the training protocols and implementation details in the original paper to conduct our experiments. We apply both the input and cost volume guidance with \ours{} following their proposed model. In Table~\ref{tab:ccvnorm}, we present the results on KITTI Depth Completion dataset~\cite{uhrig2017sparsity}. We find that the performance gain is smaller than the one in Table 2 of the main paper. We contribute it to the amount of training data, where KITTI Stereo contains about 200 pairs of data while KITTI Depth Completion is hundred times larger. Ideally, it is more likely to have large performance gains for small datasets, which highlights our framework is useful when small amount of data is available in hand.

\setlength{\tabcolsep}{0.015\linewidth}{
\begin{table}[]
    \centering
    \begin{tabular}{lcccc}
         Method & iRMSE $\downarrow$ & iMAE $\downarrow$ & RMSE $\downarrow$ & MAE $\downarrow$ \\ \hline \hline
         Wang \etal~\cite{wang20193d} & \textbf{1.40} & 0.81 & 0.7493 & 0.2525 \\
         \textbf{+ Ours} & 1.54 & \textbf{0.79} & \textbf{0.7037} & \textbf{0.2396}\\
    \end{tabular}
    \caption{\textbf{More Results of Guidance on Cost Volume.} The experiment shows the results when applying our \ours{} to the batch normalization of the cost volume.}
    \label{tab:ccvnorm}
\end{table}
}

\section{Details about the Confidence of \ours{}}
\subsection{Confidence Aggregation}
The main paper mentions that we use \textit{maximum} operation to aggregate confidence patches into the final confidence map in Equation 2 ($C'(i', j') = \max_{k \in S_k} C_k(i', j')$). The following we discuss why choosing the \textit{maximum} operation. Suppose a pixel coordinate $(i', j')$ without sparse signals ($(i', j') \neq (i_k, j_k)$, $\forall k \in S_k$) and $(i', j')$ is expanded by three nearby sparse signal sources with depth $(d_1, d_2, d_3)$ and confidence $(c_1, c_2, c_3)$, we consider two alternative aggregation operations (1) averaging the confidence and (2) interpolation with the confidence itself.

For (1) average the confidence $C(i', j') = \frac{c_1 + c_2 + c_3}{3}$, suppose the ground truth $D^*(i', j') = 50$, $(d_1, d_2, d_3) = (50, 100, 100)$, $(c_1, c_2, c_3) = (1, 0.001, 0.001)$. The nearby depth $100$ is apparently not similar to the depth $50$, so the estimated values for $c_2$ and $c_3$ are reasonable to be close to zero. Nonetheless, the values of $c_2$ and $c_3$ lower the averaged confidence to about $0.33$, which does not make sense. This case particularly happens to the occlusions or object edges.

For (2) interpolation with confidence itself $C(i', j') = \frac{c_1 \cdot c_1 + c_2 \cdot c_2 + c_3 \cdot c_3}{c_1 + c_2 + c_3}$, suppose two cases: (a) $(d^a_1, d^a_2, d^a_3) = (50, 50, 50)$, $(c^a_1, c^a_2, c^a_3) = (1, 0.01, 0.01)$ and (b) $(d^b_1, d^b_2, d^b_3) = (50, 50, 50)$, $(c^b_1, c^b_2, c^b_3) = (1, 0.9, 0.9)$. The expectation of the final confidence for case (b) should be greater than or at least equal to the final confidence for case (a), since the expanded signals in case (b) vote for higher confidence values. However, the interpolated confidence for case (b) is about $0.94$, while the one for case (a) is about $0.98$, which is opposite to the expectation.

The above two counterexamples explain why neither \textit{averaging} nor \textit{interpolation} operations are used. Our proposed \textit{maximum} operation can deal with the two cases to some degree. We look forward to some interesting and effective approaches to aggregate the confidence patches.

\subsection{Discussions on Confidence Map}
An insightful comment from one of the reviewers is that the confidence maps along the stacked axis may relate to the slanted surfaces. Suppose there are three sparse depth pixels lying on the same slanted surface (e.g., road), and a neighboring pixel on the surface is interpolated by the three pixels with confidence predicted from \ours{} network, the four pixels should form a slanted surface by projecting them to the 3D space with the intrinsic matrix. To this end, one could develop geometric constraints on the confidence from \ours{} network via the projection matrix and the assumption that neighboring points fall on the same surface. In addition, one could leverage normal visualizations to help distinguish a good confidence prediction if the slanted assumption holds. We appreciate the idea and are open to have future discussions.

\section{Impact of \oursfull{}}
Our analysis about the impact of sparse signal focuses on the following questions: (1) \emph{How much improvement comes from sparse signal guidance?} (2) \emph{How many more pixels are further improved due to the proposed \ours{} method?} and (3) \emph{are further improved pixels easy or hard cases?}
In Table~\ref{tab:changed_pixel_perc}, relatively less pixels are largely improved by comparing the ``$> 2$ d" and ``$> 0$ d" columns. Furthermore, with our method, more pixels are guided and  thus average pixel error is lower.
Finally, our method shows about $4$ times of improvement on ``$> 2$ d", which is much larger than ``$> 0$ d". This highlights that our $S^3$ can improve more on hard cases.

\setlength{\tabcolsep}{0.022\linewidth}{
\begin{table}[t]
    \centering
    \begin{tabular}{lccccc}
         \multirow{2}{*}{Method} & \multicolumn{4}{c}{\% of pixel improved} & \multirow{2}{*}{\shortstack{Avg\\Error}}\\ \cline{2-5}
         & $> 2$ d & $> 1$ d & $> 0.5$ d & $> 0$ d &  \\ \hline
         GSM & 2.4 & 6.2 & 14.6 & 88.3 & 1.370 \\
         GSM + Ours & 8.2 & 15.2 & 27.5 & 96.9 & \textbf{1.125} \\ \hline
         GDC & 0.3 & 0.9 & 2.4 & 14.7 & 0.950 \\
         GDC + Ours & 1.1 & 2.7 & 5.9 & 21.0 & \textbf{0.904} \\
    \end{tabular}
    \caption{\textbf{Impact of Sparse Signals.} With our proposed method, the same depth correction algorithm can influence more depth pixels and achieve better performance. The ``\% of pixel improved $> n$ d" denotes the percentage of pixel improved for more than $n$ disparity value owing to the depth fusion method GSM~\cite{poggi2019guided} and GDC~\cite{you2020pseudo}.
    }
    \label{tab:changed_pixel_perc}
\end{table}
}

\begin{figure}
    \centering
    \begin{subfigure}[b]{\linewidth}
        \centering
        \includegraphics[width=\linewidth]{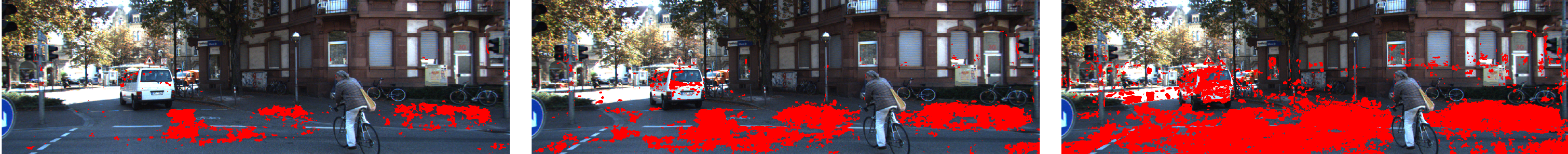}
        \caption{Guided with sparse signal.}
    \end{subfigure}
    \begin{subfigure}[b]{\linewidth}
        \centering
        \includegraphics[width=\linewidth]{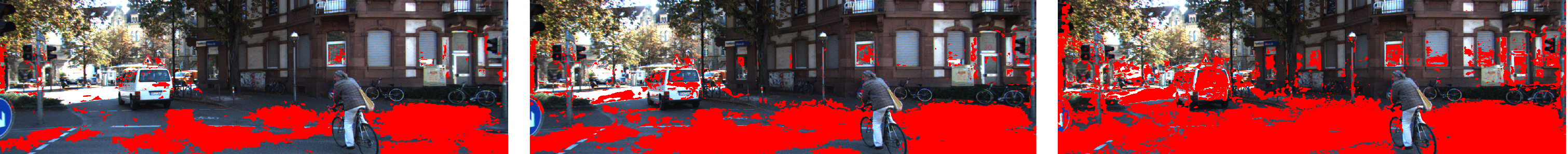}
        \caption{Guided with expanded signal.}
    \end{subfigure}
    \caption{\textbf{Impact of expansion.} Applying our expanded signal of \ours{} on GSM~\cite{poggi2019guided} can improve more depth points with the same source of sparse signal. Red points represents the pixels improved for more than $5$, $2$, and $0.5$ disparity value from left column to right, respectively. More depth points are guided with our method by comparing the two top and bottom sub-figures. Best viewed in color.}
    \label{fig:changed_perc_vis}
\end{figure}

Here we visualize an example in Figure~\ref{fig:changed_perc_vis} to show the guided pixels (red) with (b) and without (a) our method.
The region improved with the sparse signal is also improved with our method, since the expanded results of \ours{} also contains the sparse signal.
In addition, the improved and further improved region is mostly the homogeneous surface (e.g. plane road) where the stereo matching algorithm fails to find visual cues and produce accurate matches. Our method works on the homogeneous surface because the sparse signal gives the depth hint for \ours{} module to estimate the slanted information about the surface.

\section{More Visualization}
\begin{figure}
    \centering
    \includegraphics[width=\linewidth]{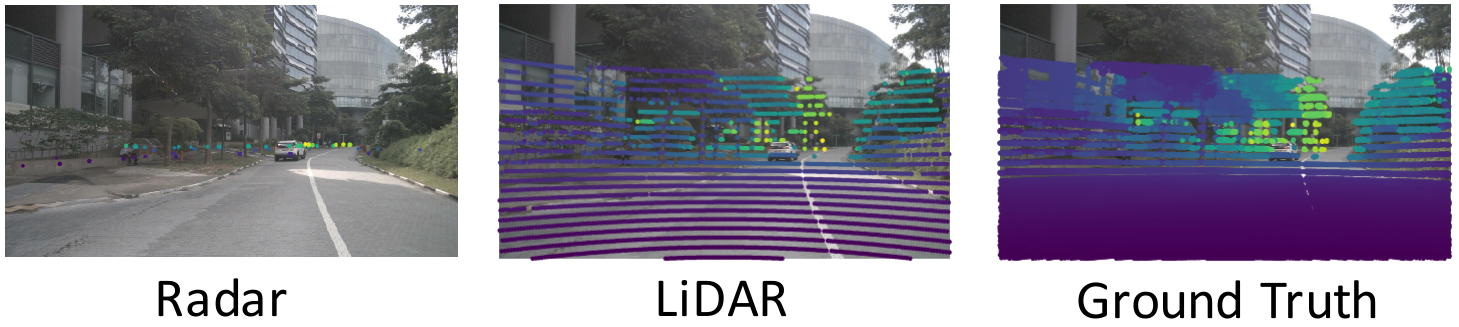}
    \caption{\textbf{Visualization of sparse signals (Radar and LiDAR) on nuScenes dataset~\cite{nuscenes2019}.}
    Images from left to right are Radar, LiDAR, and the depth ground truth accumulated from 11 nearby frames.
    The Radar and LiDAR points (visually enhanced) are extremely sparse and imbalanced.}
    \label{fig:nuScenes}
\end{figure}

We visualize the LiDAR and Radar signals with \textit{low density} and \textit{imbalanced distribution} problems in Figure~\ref{fig:nuScenes}. The elevation degree of the Radar sensor is poor so the points are mostly located at the horizontal vision line. Also, filtering operation is applied to the Radar point cloud to reduce the noise. As a result, the Radar signal is extremely sparse and imbalanced. The LiDAR signal is sparse and mostly located on the scanning lines. 

{\small
\bibliographystyle{ieee_fullname}
\bibliography{egbib}
}

\end{document}